\documentclass{article}
\usepackage{spconf,amsmath,graphicx}
\usepackage{graphicx}
\usepackage{amsmath,amsfonts,amssymb,amsthm,dsfont,mathrsfs}
\usepackage{bbm}
\usepackage{mathbbol}
\usepackage{mathtools}%
\usepackage{url}
\usepackage{cite}
\usepackage{color}
\usepackage{epstopdf}
\usepackage{dsfont}
\usepackage{ifpdf}
\usepackage{float}
\usepackage{bbm}
\usepackage{caption}
\usepackage{subcaption}

\newtheorem{definition}{Definition}

\def \bx {{\mathbf{x}}}
\def \bs {{\mathbf{s}}}
\def \bz {{\mathbf{z}}}

\def \bC {{\mathbf{C}}}

\def \bH {{\mathbf{H}}}

\def \bu {{\mathbf{u}}}

\def \mR {{\mathbb{R}}}

\def \cH {{\mathcal{H}}}
\usepackage[nodisplayskipstretch]{setspace}

\title{Predicting Brain Age using Transferable coVariance Neural Networks}
\name{Saurabh Sihag$^{\dagger}$, Gonzalo Mateos$^{\ast}$, Corey McMillan$^{\dagger}$, and Alejandro Ribeiro$^{\dagger}$}
\address{$^{\dagger}$ University of Pennsylvania, Philadelphia, PA.\\
$^{\ast}$ University of Rochester, Rochester, NY.}

\begin{document}
%

\maketitle
\begin{abstract}
The deviation between chronological age and biological age is a well-recognized biomarker associated with cognitive decline and neurodegeneration. Age-related and pathology-driven changes to brain structure are captured by various neuroimaging modalities. These datasets are characterized by high dimensionality as well as collinearity, hence applications of graph neural networks in neuroimaging research routinely use sample covariance matrices as graphs. We have recently studied covariance neural networks (VNNs) that operate on sample covariance matrices using the architecture derived from graph convolutional networks, and we showed VNNs enjoy significant advantages over traditional data analysis approaches. In this paper, we demonstrate the utility of VNNs in inferring brain age using cortical thickness data. Furthermore, our results show that VNNs exhibit multi-scale and multi-site transferability for inferring {brain age}. In the context of brain age in Alzheimer's disease (AD), our experiments show that i) VNN outputs are \emph{interpretable} as brain age predicted using VNNs is significantly elevated for AD with respect to healthy subjects for different datasets; and ii) VNNs can be \emph{transferable}, i.e., VNNs trained on one dataset can be transferred to another dataset with different dimensions without retraining for brain age prediction.

\let\thefootnote\relax\footnotetext{Submitted to 2023 IEEE International Conference on Acoustics, Speech, and Signal Processing.}

\end{abstract}
\section{Introduction}
Ageing is a biologically complex process marked by progressive anatomical and functional changes in the brain~\cite{lopez2013hallmarks}. Critically, individuals age at different rates, captured by so-called ``biological aging", and accelerated aging (e.g., when biological age is greater than chronological age) is a precursor for cognitive decline and age-related phenotypes like dementia~\cite{habes2016advanced}. Therefore, age-related biological changes are not uniform across the population. Hence, understanding the biological mechanisms of the aging brain is widely relevant because of their implications on quality of life and clinical interventions in vulnerable populations~\cite{cole2018brain}.

Different neuroimaging modalities provide complementary insights into the changes of the brain due to pathology and healthy ageing~\cite{frisoni2010clinical,chen2019functional}. Indeed, various statistical approaches to derive biomarkers of pathology from neuroimaging datasets have been widely studied in the literature~\cite{kloppel2012diagnostic}. The biomarker of interest in this paper is the \emph{brain age gap}, i.e., the difference between the predicted biological age and the chronological age. The brain age gap for a pathology can be conceptualized as a scalar representation of the accumulation of longitudinal atypical changes in the brain leading to the pathology~\cite{franke2012longitudinal}. Existing studies have reported an elevated  brain age with respect to controls for various neurodegenerative conditions, including Alzheimer's disease~\cite{franke2012longitudinal} and schizophrenia~\cite{hajek2019brain}.

Inferring brain age from different neuroimaging modalities, including structural magnetic resonance imaging (MRI), functional MRI, and positron emission tomography, has been an active area of research~\cite{lee2022deep, yin2022deep,millar2022multimodal,dafflon2020automated,baecker2021brain,aycheh2018biological}. The statistical approaches utilized for predicting brain age are usually specialized to the input structure (for e.g., convolutional neural networks for raw MRI images~\cite{lee2022deep, yin2022deep}). In this paper, we aim to predict brain age using cortical thickness measures derived from structural MRI images. The cortical thickness measures correspond to different brain regions that are pre-defined according to a brain atlas used in the processing of the MRI images~\cite{de2013parcellation}. Cortical thickness measures evolve with normal ageing~\cite{mcginnis2011age} and are affected due to neurodegeneration~\cite{hayes2017mild}. Thus, the age-related and disease severity related variations also appear in anatomical covariance matrices evaluated from the correlation among the cortical thickness measures across a population~\cite{evans2013networks}. 

Neuroimaging data analyses face challenges due to limited sample size, high cost of data acquisition, and heterogeneity in datasets due to induced inter-site and inter-scanner variability~\cite{zhu2011quantification}. To tackle such challenges, existing studies deploy techniques from meta learning and transfer learning to develop data-efficient and compatible algorithms~\cite{ardalan2022transfer,yang2022data}. Specifically, transfer learning is closely related to domain adaptation, where the inference algorithm trained on a source domain is fine tuned for performance in a target domain~\cite{ardalan2022transfer}. In~\cite{yang2022data}, learnable brain atlas transformations are studied to combine datasets organized according to different brain atlases.

In our recent work, we have studied coVariance neural networks (VNNs) that derive their architecture from graph convolutional networks~\cite{ruiz2021graph} and operate on the sample covariance matrix as a graph~\cite{sihag2022covariance}. VNNs have two key features: i) VNN outputs are~\emph{stable} to perturbations in the covariance matrices; and ii) VNNs can be~\emph{transferable} to datasets of dimensions different from that of the training set while preserving performance. Both these features translate into significant advantages in terms of generalizability and reproducibility over traditional data analysis approaches like principal component analysis and regression that are typically used for the analysis of cortical thickness data~\cite{evans2013networks, baecker2021brain,aycheh2018biological}. The notion of transferability in this paper is derived from the properties of coVariance filters whose parameters do not depend on the covariance matrix and therefore, the VNNs can be transferred to a dataset of a different dimension. Hence, the notion and implementation of transferability in this paper is distinct from the approaches in~\cite{ardalan2022transfer} and~\cite{yang2022data}. Our contributions in this paper are summarized below. 

\noindent
\textbf{Contributions}: We study a VNN-based framework for brain age prediction on various datasets consisting of controls, subjects with mild cognitive impairment (MCI) and subjects with Alzheimer's disease (AD), where MCI is an intermediate clinical stage between no cognitive impairment and AD~\cite{bennett2005mild}. We consider two independent datasets with cortical thickness features extracted according to distinct brain atlases or templates. One dataset is a multi-scale cortical thickness dataset curated according to different resolutions of Schaefer parcellation~\cite{schaefer2018local}. In this context, our observations are as follows:

\noindent
1. {\emph{Interpretablity}}: Brain age predicted using VNNs is significantly elevated for AD with respect to controls and the brain age for the MCI cohort is intermediate between AD and controls. This observation is qualitatively consistent with the findings of other brain age approaches in the literature~\cite{millar2022multimodal}. Furthermore, our results show that brain age gap correlates significantly with clinical dementia rating (CDR), which is a marker of dementia severity~\cite{morris1991clinical}. These findings imply that brain age predicted using VNN is a potential biomarker for AD and clinically interpretable.

\noindent
2. {\emph{Transferability}}: The proposed brain age prediction framework exhibits multi-scale transferability, i.e., it can be transferred across the different scales of the multi-scale dataset with no retraining and minimal deviation in performance. Moreover, in contrast to existing frameworks that rely on atlas transformation for transfer learning or domain adaptation between neuroimaging datasets~\cite{yang2022data}, we show that the VNN-based framework is transferable for brain age prediction while preserving interpretability without any learnable atlas transformation.

\section{Setting}\label{Set}
In this section, we briefly discuss the preliminaries of the cortical thickness datasets. We denote the number of subjects in the dataset by $n$. For every subject, we have mean cortical thickness data available at $m$ distinct cortical regions. Therefore, for a subject ${i \in \{1,\dots,n\}}$, we have a vector of mean cortical thickness data ${\bx_i\in \mR^{m\times 1}}$ and their chronological age $y_i\in \mR$. Using the samples $\{\bx_i\}_{i=1}^n$, the covariance matrix is evaluated as
\begin{align}\label{cov}
    \bC \triangleq \frac{1}{n} \sum\limits_{i=1}^n (\bx_i - \bar\bx)(\bx_i - \bar\bx)^{\sf T}\;,
\end{align}
where $\bar\bx$ is the sample mean of $n$ samples and $\cdot^{\sf T}$ is the transpose operator. Furthermore, we also investigate brain age prediction on a multi-scale cortical thickness dataset described next.

\noindent
\textbf{Multi-scale Cortical Thickness Dataset}: In a multi-scale setting, the cortical thickness data across the whole brain is available at different scales or resolutions. Examples of brain atlases that allow multi-scale brain parcellations include Schaefer's atlas~\cite{schaefer2018local} and Lausanne atlas~\cite{hagmann2008mapping}. A multi-scale brain atlas partitions the cortical surface into variable number of regions at different scales. For a multi-scale dataset with $S > 1$ scales, let the dimension of cortical thickness at $s$-th scale be denoted by $m_s, s \in \{1,\dots,S\}$. In the $s$-th scale, the cortical thickness data for subject $i$ is given by ${\bx_i^{m_s} \in \mathbb{R}^{m_s \times 1}}$. Subsequently, we denote the covariance matrix at scale $m_s$ by $\bC_{m_s}$, where $\bC_{m_s}$ is evaluated from $\bx_i^{m_s}, i\in \{1,\dots,n\}$ according to~\eqref{cov}. Therefore, a multi-scale cortical thickness dataset from $n$ subjects is represented as $\{\{\bx_i^{m_s}\}_{i=1}^n\}_{s=1}^S$. Next, we discuss the framework for brain age prediction using VNNs.

\section{Methods}\label{method}
We start by discussing the architecture and the notion of multi-scale transferability of VNNs.
\subsection{coVariance Neural Networks}
VNNs inherit the architecture of graph convolutional networks and therefore, consist of a bank of convolutional filters and pointwise non-linear activation function. The convolutional filters for VNNs, referred to as coVariance filters, operate on the sample covariance matrix. If $m$ dimensions of the data can be considered as individual nodes of an undirected graph, the covariance matrix $\bC$ represents its adjacency matrix, where the off-diagonal elements of $\bC$ represent the linear relationship between different dimensions of the dataset. 
\subsubsection{Architecture}
Analogously to a graph convolutional filter that combines information according to the graph topology, we define coVariance filter as
\begin{align}\label{fil}
    \bH(\bC) \triangleq \sum\limits_{k=0}^K h_k \bC^k \;,
\end{align}
where parameters $\{h_k\}_{k=0}^K$ are scalars and referred to as filter taps. Accordingly, the output of the graph filter for some data $\bx \in \mathbb{R}^{m\times 1}$ is given by $\bH(\bC)\bx = \sum_{k=0}^K h_k \bC^k \bx$. In a single layer of VNN, the coVariance filter output is further passed through a non-linear activation function (e.g. ${\sf ReLU}, \tanh$), such that, the output of a single layer VNN with input $\bx$ is given by~$\bz = \sigma\Big(\bH(\bC) \bx\Big)$,
where $\sigma(\cdot)$ is a pointwise non-linear function, such that, $\sigma(\bu) = [\sigma(u_1), \dots, \sigma(u_m)]$ for $\bu = [u_1, \dots, u_m]$.

To incorporate sufficient expressive power in the VNN architecture for a learning task, in practice, a VNN may consist of multiple layers and parallel features per layer to form filter banks at every layer. For a VNN layer with $F_{\sf in}$ features at input, such that, $\bx_{\sf in} = [\bx_{\sf in}[1],\dots,\bx_{\sf in}[F_{\sf in}]]$, and $F_{\sf out}$ features at the output, such that, $\bx_{\sf out} = [\bx_{\sf out}[1],\dots,\bx_{\sf out}[F_{out}]]$, the relationship between the $f$-th output $\bx_{\sf out}[f]$ and the input $\bx_{\sf in}$ is given by
\begin{align}
   \bx_{\sf out}[f] &=  \sigma\left(\sum\limits_{g = 1}^{F_{\sf in} } \bH_{fg}(\bC)\bx_{\sf in} [g] \right)\label{vfbnk}\;.
\end{align}
Finally, the node-level features at the output of the final VNN layer are aggregated via unweighted mean, thus, rendering the VNN architecture to be~\emph{permutation invariant}. We denote the VNN architecture using the notation $\Phi(\bx;\bC,\cH)$, where $\cH$ is the set of all filter taps associated with the coVariance filters in all layers of the VNN.
For a supervised learning objective with a training dataset $\{\bx_i, y_i\}_{i=1}^n$, the filter taps in $\cH$ are chosen to minimize the training loss, i.e.,
\begin{align}
    \cH_{\sf opt} = \min_{\cH} \frac{1}{n}\sum\limits_{i=1}^n \ell(\Phi(\bx_i;\bC,\cH), y_i)\;,
\end{align}
where $\ell(\cdot)$ is the mean squared error (MSE) loss function that satisfies ${\ell(\Phi(\bx_i;\bC,\cH), y_i) = 0}$ iff $\Phi(\bx_i;\bC,\cH)= y_i$. 
\subsubsection{Transferability of VNNs}

The filter taps in the graph filter in~\eqref{fil} are independent of the dimension of the covariance matrix. Therefore, in the multi-scale setting, we can readily replace the covariance matrix at scale $m_1$, i.e., $\bC_{m_1}$ in $\bH(\bC_{m_1})$ with $\bC_{m_2}$ to process the cortical thickness data at scale $m_2$ using VNN trained at scale $m_1$. The VNN with filter taps ${\cal H}$ is transferable in the multi-scale setting if the difference between the outputs $\Phi(\bx_i^{m_1}; \bC_{m_1}, {\cal H})$ and $\Phi(\bx_i^{m_2}; \bC_{m_2}, {\cal H})$ for subject $i$ is bounded. This notion is formalized in the following definition.
\begin{definition}[Transferable coVariance Neural Networks]
Consider a multi-scale dataset $\{\bx_i^{m_1},\bx_i^{m_2}\}_{i=1}^n$, where $\bx_i^{m_1} \in \mR^{m_1\times 1}$ and $\bx_i^{m_2}\in \mR^{m_2\times 1}$. 
A VNN architecture $\Phi(\cdot)$ with a set of filter taps $\cH$ is  transferable between data $\{\bx_i^{m_1}\}_{i=1}^n$ and $\{\bx_i^{m_2}\}_{i=1}^n$ if we have
\begin{align}\label{transfer}
    |\Phi(\bx_i^{m_1};\bC_{m_1}, \cH) - \Phi(\bx_i^{m_2};\bC_{m_2},\cH)| \leq \varepsilon\;,
\end{align}
where $\varepsilon > 0$ is some finite constant, $\bC_{m_1}$ is the sample covariance matrix for data $\{\bx_i\}_{i=1}^n$ and $\bC_{m_2}$ is the sample covariance matrix for data $\{\bs_i\}_{i=1}^n$. 
\end{definition}
\noindent Existing studies report the transferability property to hold for GNNs between graphs that are sampled from the same limit object (graphon)~\cite{ruiz2020graphon}. Since the multi-scale dataset (described in Section~\ref{Set}) involves a variable number of partitions of the same surface across scales, we anticipate VNNs to be transferable for applications involving such datasets. The implication of the permutation-invariance of VNNs is that the cortical thickness datasets curated according to different brain atlases or templates do not have to be aligned for~\eqref{transfer} to hold. Therefore, we will also investigate whether transferability exists for datasets of different dimensions collected from independent sites. Moreover, the property of transferability  enables VNNs to be computationally efficient (for instance, by training on low resolution data and deploying on high resolution data)~\cite{sihag2022covariance}.  


In this paper, we deploy VNN as a regression model to learn the relationship between cortical thickness and chronological age, thus, capturing the variations in cortical thickness due to ageing. The details on the evaluation of brain age and using it as a biomarker of pathology are discussed next.
\subsection{Brain Age Evaluation and $\Delta$-Age}
The gap between estimated brain age and the chronological age is a biomarker of pathology~\cite{franke2012longitudinal}. Using VNNs, we obtain an estimate $\Phi(\bx_i;\bC,\cH)$ for chronological age $y_i$. However, there may exist a systemic bias in the gap between $\Phi(\bx_i;\bC,\cH)$ and $y_i$, where the age is underestimated for older subjects and overestimated for younger subjects~\cite{beheshti2019bias}. Such a bias may exist, for example, when the correlation between VNN output and chronological age is smaller than~1. In practice, this age bias can confound the interpretations of brain age. Therefore, to correct for age-related bias, we adopt a linear model based post hoc bias correction approach from the existing literature~\cite{de2020commentary}. Under this approach, we follow the following bias correction steps on the VNN estimated age $\Phi(\bx;\bC,\cH)$ to obtain the brain age $\hat y_{\sf B}$ for a subject with chronological age $y$ and cortical thickness data~$\bx$:

\noindent
{\bf Step 1:} Fit a linear regression model to determine $\alpha$ and $\beta$ in the following model:
\begin{align}\label{s1}
     \Phi(\bx;\bC,\cH) - y = \alpha y + \beta\;. 
\end{align}
\noindent
{\bf Step 2:} Obtain corrected predicted age or brain age as follows:
\begin{align}\label{s2}
   \hat y_{\sf B} = \Phi(\bx;\bC,\cH) - (\alpha y + \beta)\;. 
\end{align}
Therefore, $\hat y_{\sf B}$ forms the brain age after bias correction for a subject with chronological age $y$. The gap between $\hat y_{\sf B}$ and $y$ is the biomarker of interest which is defined below. 
\begin{definition}[$\Delta$-Age] For a subject with cortical thickness $\bx$ and chronological age $y$, the brain age gap is defined as
\begin{align}
     \Delta\text{-Age}\triangleq \hat y_{\sf B} - y\;,
\end{align}
where $\hat y_{\sf B}$ is determined from the VNN estimate $\Phi(\bx;\bC,\cH)$ and $y$ according to steps in~\eqref{s1} and~\eqref{s2}.
\end{definition}
\noindent
In our experiments, we obtain VNN estimates on the combined dataset of healthy subjects and pathological subjects. The age-bias correction in~\eqref{s1} and~\eqref{s2} is performed over healthy subjects to account for bias in the VNN estimates due to healthy ageing and applied to the pathological cohort. Further, the distributions of $\Delta$-Age are obtained for healthy controls and pathological subjects. $\Delta$-Age for subjects with pathology is expected to be elevated as compared to healthy subjects.

\section{Results}
In this section, we discuss our experiments on brain age prediction for different datasets using the methodology outlined in Section~\ref{method}. The datasets consist of heterogeneous populations of healthy subjects (HC), subjects with MCI, and subjects with AD.  We consider the following datasets:

\noindent
{\bf Multi-scale FTDC Datasets:} These datasets consist of the cortical thickness data extracted at different resolutions from healthy subjects (HC; $n = 170$, age = $64.26 \pm 8.26$ years, 101 females), subjects with MCI (MCI; $n = 53$, age = $68.56 \pm 8.58$ years, 22 females), and subjects with AD (AD; $n = 62$, age = $67.25 \pm 8.83$ years, 30 females). For each subject, we have the cortical thickness data curated according to multi-resolution Schaefer atlas~\cite{schaefer2018local}, at 100 parcel, 300 parcel, and 500 parcel resolutions. 
Accordingly, we form three datasets: FTDC100, FTDC300 and FTDC500, which form the cortical thickness datasets corresponding to 100, 300 and 500 features resolutions, respectively. For $23$ MCI and $17$ AD subjects, we also have the CDR sum of boxes scores. CDR sum of boxes scores are commonly used in clinical and research settings to quantify dementia severity. A higher CDR score is associated with cognitive and functional decline~\cite{o2008staging}.

\noindent 
{\bf ABC Dataset:} This dataset is collected from a population of $308$ subjects, which is composed of healthy subjects (HC; $n= 171$ subjects, age = $71.88 \pm 7.05$ years,  116 females), subjects with MCI (MCI; $n= 47$, age = $72.85 \pm 8.52$ years, 18 females), subjects with AD (AD; $n = 51$, age = $70.92 \pm 8.23$ years, 23 females), and remaining subjects with other forms of dementia. For each
subject, joint-label fusion~\cite{wang2012multi} was used to quantify mean cortical thickness in $m = 96$ cortical regions. 

Our goal is to evaluate whether the VNN-based brain age prediction framework outputs can act as biomarkers of disease in AD or MCI. For this purpose, we focus on group differences in $\Delta$-Age for HC, MCI, and AD for all experiments and the correlation of $\Delta$-Age with CDR scores. To determine the success of transferability, we evaluate if group differences in $\Delta$-Age are retained and if the $\Delta$-Age scores correlate with CDR scores after transferring the model. In Section~\ref{age1}, we provide the experimental details and results for the brain age prediction task on FTDC300 and ABC datasets. The performances reported in this section correspond to the training performance. In Section~\ref{multiscale}, we leverage the transferability of VNNs to test the brain age prediction models from Section~\ref{age1} for multi-scale and multi-site transferability. 
\subsection{Brain Age Prediction using VNN}\label{age1}
We first evaluate the proposed VNN-based framework for brain age prediction on FTDC300 and ABC datasets. The VNN models are trained using two distinct approaches for FTDC300 and ABC datasets. For FTDC300, we train VNNs to predict chronological age using only HC cohort. In contrast, for ABC dataset, we use the complete, hetereogeneous dataset for training VNNs. The former approach is more commonly used, while the latter has been used to disentangle neurodegeneration from healthy ageing~\cite{hwang2022disentangling}. The subsequent steps to evaluate the brain age for HC, MCI, and AD cohorts are the same in both approaches.

\noindent
\textbf{FTDC300 Dataset:} The VNN is trained as a regression model between cortical thickness and chronological age using data from only HC cohort. For this purpose, we randomly divide the HC cohort in a $80/10/10$ train/validation/test split. The sample covariance matrix is evaluated from the training set and normalized, such that its largest eigenvalue is $1$. For FTDC300, the VNN model consists of $2$-layer architecture, with $2$ filter taps per layer, $44$ features per dimension for $m = 300$ dimensions in both layers, and is trained using mean squared error as the loss function. The learning rate for the Adam optimizer is $0.0033$. The hyperparameters for the VNN architecture are chosen using a hyperparameter optimization framework applied on the training set~\cite{akiba2019optuna}. The model is trained up to $100$ epochs and the best model is chosen based on its performance on the validation set. We train $100$ VNN models on different permutations of the training and validation sets. 

To evaluate brain age for the FTDC300, we use the sample covariance matrix from the whole dataset and evaluate the VNN predicted age for a subject as the mean of all $100$ VNN models. The bias due to chronological age is corrected according to the methodology described in~\eqref{s1} and~\eqref{s2}, where the linear regression model for bias correction is trained on the HC cohort. The MAE between the predicted brain age (after bias correction of VNN outputs) and chronological age for HC subjects is $3$ years, MCI subjects is $5.25$ years and AD subjects is $7.06$ years. Figure~\ref{transfer_fig1} depicts the distributions of $\Delta$-Age for HC, MCI, and AD cohorts for the FTDC300 dataset. Significant group differences exist among the $\Delta$-Age for AD, MCI, and HC (ANOVA: $F$-value = 35.92, $p$-value$<10^{-10}$). As expected, the $\Delta$-Age for HC is significantly lower than that for AD cohort (post hoc Tukey's HSD test: FWER corrected $p$-value = 0.001 for AD vs HC), with the $\Delta$-Age for MCI intermediate between the~two (post hoc Tukey's HSD test: FWER corrected $p$-values are 0.002 for MCI vs HC and 0.001 for AD vs MCI).

\noindent
\textbf{ABC Dataset:} For this dataset, we use cortical thickness data~\emph{for the complete dataset} for training VNNs. The dataset is split into an $80/10/10$ train/validation/test split. The VNN architecture consists of $2$ layers with $2$ filter-taps per layer, $13$ features per dimension for $m=96$ dimensions in both layers. The selection of VNN hyperparameters and the training procedure are similar to that for FTDC300 dataset. 
The subsequent brain age evaluation procedure is similar to that followed for FTDC300 and focuses on correcting for bias due to healthy ageing. The MAE between predicted brain age and chronological age is $4.09$ years for HC, $6.8$ years for MCI, and $8.23$ years for AD. The box plots illustrating the distributions of $\Delta$-Age for the three cohorts in the ABC dataset are shown in Fig.~\ref{transfer_fig2}. Significant group differences exist between $\Delta$-Age for the AD, MCI, and HC cohort (ANOVA: $F$-value= 35.83, $p$-value$<10^{-10}$). The $\Delta$-Age for HC is significantly lower than that for AD (post hoc Tukey's HSD test: FWER corrected $p$-value = 0.001 for AD vs HC), with MCI intermediate between the~two (post hoc Tukey's HSD test: FWER corrected $p$-values are 0.001 for AD vs MCI and 0.0762 for MCI vs HC). Therefore, the approaches adopted for FTDC300 and VNN datasets provide similar qualitative results in terms of relative elevation of brain age on the HC-MCI-AD spectrum. 

\subsection{Transferability of VNNs for Brain Age Prediction}\label{multiscale}
In this set of experiments, we evaluate the transferability of brain age prediction frameworks trained on FTDC300 and ABC datasets. We focus on two aspects of transferability: i) multi-scale transferability from FTDC300 to FTDC100 and FTDC500 datasets; and ii) multi-site transferability from FTDC300 to ABC dataset and ABC dataset to FTDC100, FTDC300, and FTDC500 datasets. 

\begin{figure}[t]
  \centering
  \includegraphics[scale=0.2]{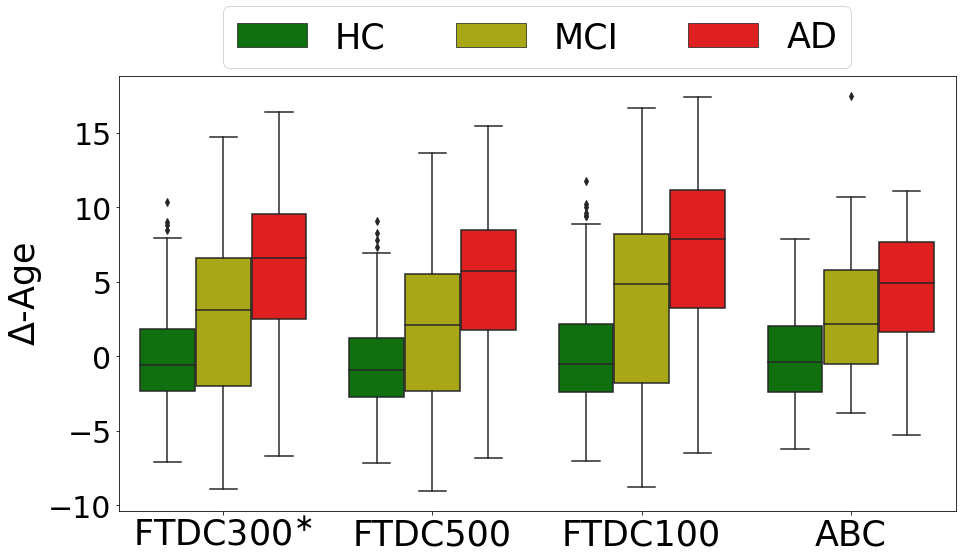}
   \caption{$\Delta$-Age for all datasets with FTDC300 dataset as training set.}
   \label{transfer_fig1}
\end{figure}

\noindent
\textbf{Multi-Scale Transferability}: 
Our results in~\cite[Section 5.2]{sihag2022covariance} show that VNN outputs are transferable as per Definition~1 (in terms of MAE) across the multi-scale FTDC datasets. Therefore, we transfer the filter taps from the VNN models trained on FTDC300 and the age bias-correction linear model to determine the brain age using cortical thickness from FTDC100 (100 features resolution) and FTDC500 (500 features resolution). Our results in Figure~\ref{transfer_fig1} show that the significant group differences exist in $\Delta$-Age for the three cohorts in both FTDC100 and FTDC500 (ANOVA: $p$-value$<10^{-10}$), and post hoc analyses reveals that $\Delta$-Age for AD is significantly elevated with respect to HC, with MCI intermediate between the two for FTDC100 and FTDC500 datasets. Thus, these findings are qualitatively consistent with those on FTDC300 model and hence, the brain age prediction model learnt on FTDC300 can be transferred to FTDC100 and FTDC500 for brain age prediction. 

\begin{figure}[t]
  \centering
  \includegraphics[scale=0.2]{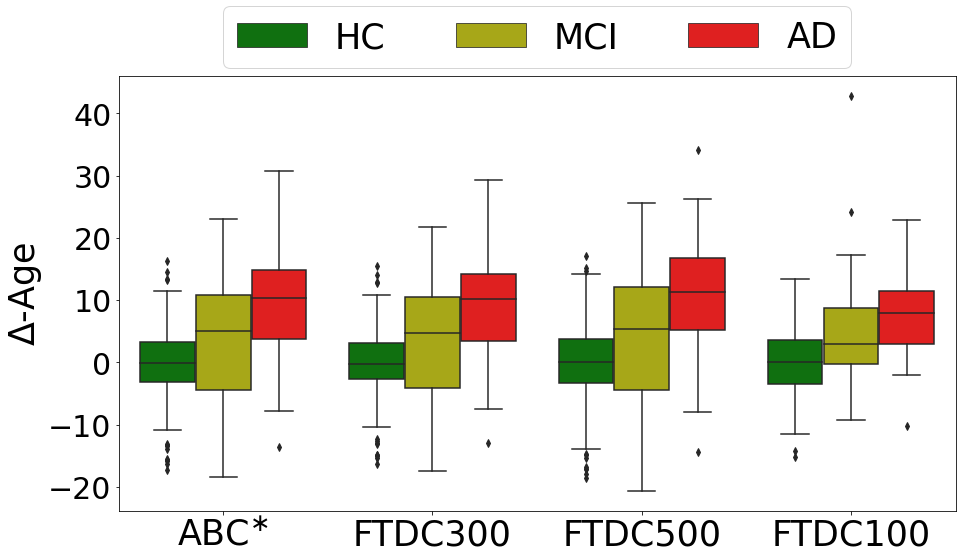}
   \caption{$\Delta$-Age for all datasets with ABC dataset as training set.}
   \label{transfer_fig2}
\end{figure}

\noindent
\textbf{Multi-Site Transferability}\label{multisite}:
We evaluate the multi-site transferability from FTDC300 to ABC, and ABC to FTDC100, FTDC300, and FTDC500 datasets.  The methodology to evaluate multi-scale transferability is similar to that for multi-site transferability with one difference: here, we re-evaluate the linear model for age bias correction on the HC cohort in the test dataset. We remark that the relative elevation of $\Delta$-Age for AD and MCI with respect to HC is preserved even without re-learning the age bias correction model in the test set. However, in this scenario, there exists a constant offset in the $\Delta$-Age for HC when the model is transferred from ABC dataset to FTDC dataset or FTDC300 to ABC dataset. In Figures~\ref{transfer_fig1} and~\ref{transfer_fig2}, we observe that the VNN models learnt on FTDC300 and ABC datasets using the procedures described in Section~\ref{age1} can be transferred to ABC and FTDC datasets, respectively, while preserving interpretability of $\Delta$-Age. Specifically, we observe the pattern of progressive increase in $\Delta$-Age from HC to MCI to AD after transferring the VNNs to the test datasets. In the experiment to evaluate transferability from FTDC300 to ABC dataset (Fig.~\ref{transfer_fig1}), we observe significant group differences in $\Delta$-Age (ANOVA: $F$-value = 43.15) and significant pairwise differences (post hoc Tukey's HSD test: $p$-values $<0.05$ for all pairwise comparisons). For case of ABC to FTDC300 (Fig.~\ref{transfer_fig2}), we make similar observations: $F$-value = 31.76 for ANOVA and $p$-values $<0.05$ for all post hoc tests. Therefore, we observe evidence of multi-site transferability for brain age prediction.

\subsection{$\Delta$-Age and CDR Sum of Boxes Scores} 
The $\Delta$-Age inferred from the brain age prediction model trained on the FTDC300 dataset is significantly associated with CDR sum of boxes scores from $40$ MCI and AD subjects ($\rho = 0.38$, $p$-val $= 0.013$). Furthermore, the $\Delta$-Age inferred for the FTDC300 datset from the prediction model trained on the ABC dataset is also significantly associated with the CDR sum of boxes scores  ($\rho = 0.44$, $p$-val $= 0.003$). The observations in this section and Section~4.2 imply that the $\Delta$-Age derived using our framework is clinically interpretable and the interpretability is retained after transferability.


\section{Conclusions}
In this paper, we have proposed a VNN-based brain age predictor from cortical thickness data. By leveraging the property of transferability in VNNs, we have illustrated cross-resolution and cross-site transferability of brain age prediction. Moreover, in the context of Alzheimer's disease, our experiments have shown that the brain age predictions are biomarkers for AD and clinically interpretable. Future work includes expanding the application to other pathologies and neuroimaging modalities. 

\iftrue
\section{Acknowledgements}
The ABC dataset was provided by the Penn Alzheimer’s Disease Research Center
(ADRC; NIH AG072979) at University of Pennsylvania. The MRI data for FTDC datasets were provided by the Penn Frontotemporal Degeneration Center (NIH AG066597). Cortical thickness data
were made available by Penn Image Computing and Science Lab at University of Pennsylvania.

\section{Data Availability}\label{dataavail}
Data for experiments on ABC and FTDC datasets may be requested through \url{https://www.pennbindlab.com/data-sharing} and upon review by the University of Pennsylvania Neurodegenerative Data Sharing Committee, access will be granted upon reasonable request.
\fi
\bibliographystyle{IEEEbib}
{
\bibliography{icassp2023}}

\end{document}